\PassOptionsToPackage{main=english,russian}{babel}
\PassOptionsToPackage{T2A}{fontenc}
\documentclass[onecolumn]{misraj}
\usepackage{lipsum}
\usepackage{pdfpages}
\usepackage{colortbl}
\usepackage{tabulary}
\usepackage{longtable}
\usepackage{array}
\usepackage{placeins}
\usepackage{tablefootnote}
\usepackage{tcolorbox}
\usepackage{wrapfig}
\usepackage{adjustbox}
\usepackage{float}
\usepackage{caption}
\usepackage{breqn} % For multiline equations
\usepackage[toc,page]{appendix}
\usepackage{pifont}
\usepackage{multirow}
\usepackage{tabularx}

\usepackage[utf8]{inputenc}

\usepackage{arydshln}

\usepackage{xspace} % 
\usepackage{makecell} % For line breaks in table cells
\usepackage{multirow}

\usepackage[framemethod=TikZ]{mdframed}
\usepackage{tcolorbox}
\usepackage{multicol}

\usepackage[utf8]{inputenc}
\usepackage[T1]{fontenc}
\usepackage[english]{babel} % don't use [arabic] here with arabtex
\usepackage{longtable}
\usepackage{arabtex}
\usepackage{utf8}
\setcode{utf8}

\definecolor{lightblue}{RGB}{211, 227, 252} % Light blue => datacard
\definecolor{bgblue}{RGB}{247, 250, 255} % datacard background

\newcommand*\colourcheck[1]{%
  \expandafter\newcommand\csname #1check\endcsname{\textcolor{#1}{\ding{52}}}%
}
\newcommand*\colourcross[1]{%
  \expandafter\newcommand\csname #1cross\endcsname{\textcolor{#1}{\ding{55}}}%
}
\DeclareSymbolFont{extraup}{U}{zavm}{m}{n}
\DeclareMathSymbol{\vardiamond}{\mathalpha}{extraup}{87}
\colourcheck{blue}
\colourcheck{green}
\colourcheck{red}

\colourcross{blue}
\colourcross{green}
\colourcross{red}
\usepackage{nicematrix}
\usepackage{comment}
\usepackage{amssymb}
\usepackage{arabtex}
\usepackage[utf8]{inputenc}
\usepackage[bottom]{footmisc}
\usepackage{svg}
\usepackage{spverbatim}
\RequirePackage[hidelinks,colorlinks=true,linkcolor=blue,citecolor=blue]{hyperref}
\setcitestyle{number,square}
\definecolor{deeppurple}{HTML}{9e02f7}
\definecolor{forestgreen}{HTML}{2e7d43}

\hypersetup{urlcolor=deeppurple}

\title{Wasm: A Pipeline for Constructing Structured Arabic Interleaved Multimodal Corpora}

\multiauthors
\author{
    name={Khalil Hennara},
    email={hennara@misraj.ai}
}
\author{
    name={Ahmad Bastati},
    email={bastati@misraj.ai}
}
\author{
    name={Muhammad Hreden},  
    email={hreden@gmail.com}
}
\author{
    name={Mohamed Motasim Hamed},
    email={hamed@misraj.ai}
}
\author{
    name={Zeina Aldallal },
    email={aldallal@misraj.ai}
}
\author{
    name={Sara Chrouf},
    email={sara.chrouf@misraj.ai}
}
\author{
    name={Safwan AlModhayan},
    email={safwan@misraj.ai}
}

\affiliations{
 \includegraphics[width=2cm]{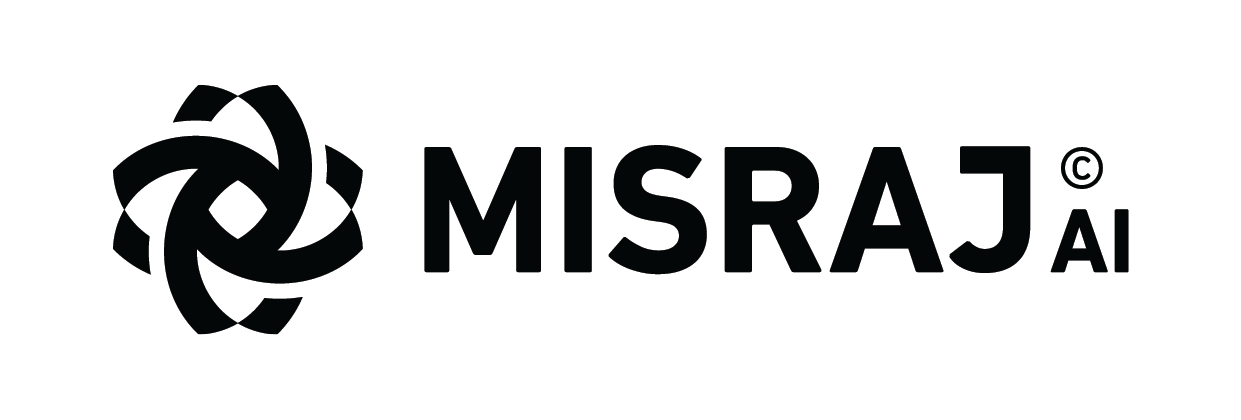}\\
     Khobar, Saudi Arabia \\
     {hennara,bastati,hreden,hamed,aldallal,chrouf,safwan}@misraj.ai \\
}
\date{\today}
\abstract{
The performance of large language models (LLMs) and large multimodal models (LMMs) depends heavily on the quality and scale of their pre-training datasets. Recent research shows that large multimodal models trained on natural documents where images and text are interleaved outperform those trained only on image–text pairs across a wide range of benchmarks, leveraging advanced pre-trained models to enforce semantic alignment, image-sequence consistency, and textual coherence. For Arabic, however, the lack of high-quality multimodal datasets that preserve document structure has limited progress. In this paper, we present our pipeline \textit{\textbf{Wasm}\textsuperscript{†}} for processing the Common Crawl\textsuperscript{*} dataset to create a new Arabic multimodal dataset that uniquely provides markdown output. Unlike existing Arabic corpora that focus solely on text extraction, our approach preserves the structural integrity of web content while maintaining flexibility for both text-only and multimodal pre-training scenarios. We provide a comprehensive comparative analysis of our data processing pipeline against those used for major existing datasets, highlighting the convergences in filtering strategies and justifying our specific design choices. To support future research, we publicly release a representative dataset dump along with the multimodal processing pipeline for Arabic.
}

\begin{document}

\renewcommand{\thefootnote}{\fnsymbol{footnote}}
\footnotetext[2]{\textbf{Wasm}: meaning 'tag' or'mark,' which reflects the unique preservation of web markup structures in our data set.}
\footnotetext[1]{\url{https://commoncrawl.org}}
\renewcommand{\thefootnote}{\arabic{footnote}}
\setcounter{footnote}{0}

\section{Introduction}
\label{sec:introduction}

The performance of Large Language Models (LLMs) is fundamentally tied to the quality and scale of their training data. While the Web offers a vast repository of text, raw web scrapes are rife with noise, including advertisements, low-quality text, and formatting artifacts. Consequently, the development of robust data processing pipelines has become a critical area of research. This is particularly true for non-English languages like Arabic, where high-quality, large-scale corpora are less common.  Moreover, existing Arabic datasets typically emphasize plain text extraction, discarding valuable structural cues (e.g., document layout, formatting, and image associations) that are crucial for training multimodal models.

Recent studies emphasize that interleaved image–text data, which preserves the natural sequence of textual and visual elements within documents, is essential for training advanced multimodal models~\cite{mmc4,OBELICS,omnicorpus,mint1T,comm}. Unlike isolated caption–image pairs, interleaved corpora capture document-level structure, allowing models to learn long-range dependencies, maintain narrative coherence, and align images with text across multiple segments or pages. This structure enables richer multimodal reasoning, such as following temporal or story-driven progressions, comparing various images within a context, and establishing complex instructions in visual evidence, while also enhancing the model's ability to generate coherent, interleaved outputs.

In contrast, existing Arabic resources typically emphasize plain text extraction, discarding valuable structural cues such as document layout, formatting, and image associations, precisely the elements that interleaved corpora preserve and leverage for training advanced multimodal models.

\begin{figure}[ht]
  \centering
  \includegraphics[width=0.8\linewidth]{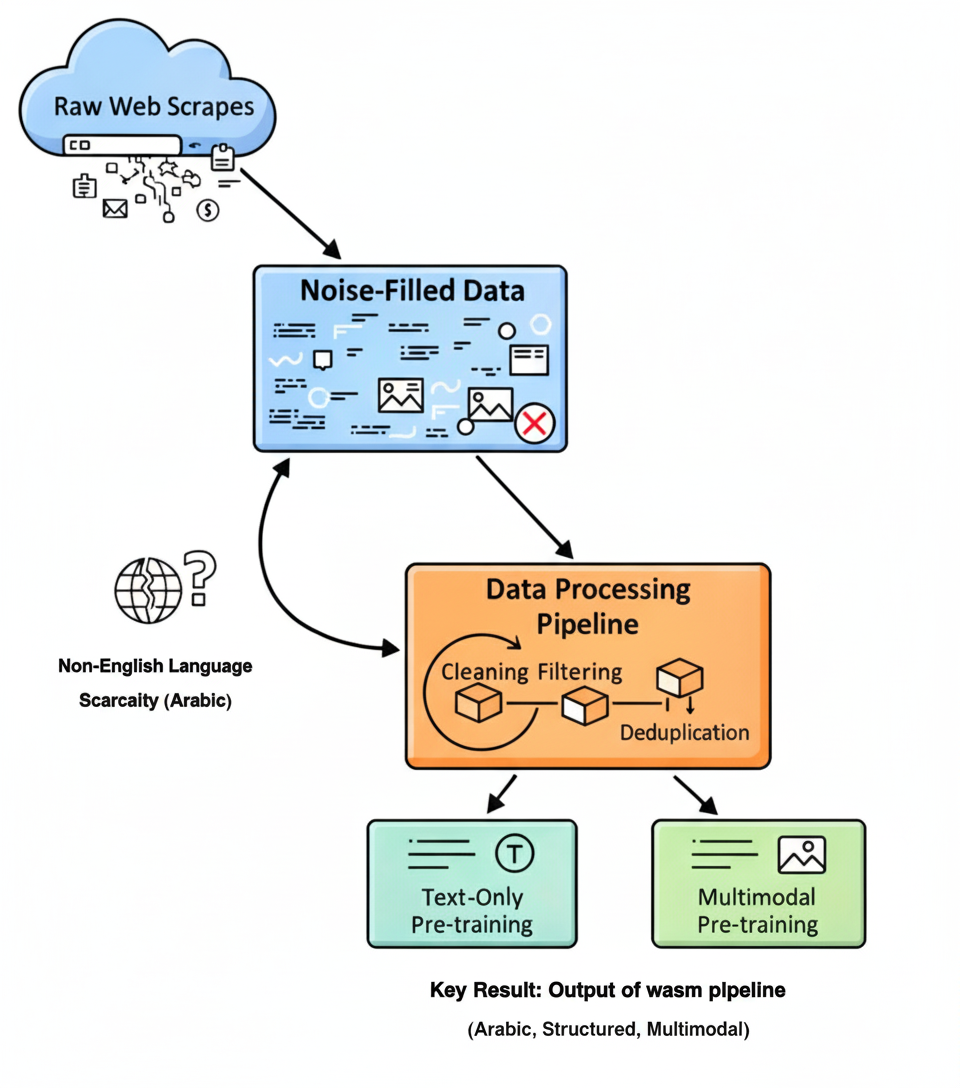} % adjust path
  \caption{Overview of our data processing pipeline.}
  \label{fig:pipeline}
\end{figure}

\begin{figure}[ht]
    \centering
    \includegraphics[width=\linewidth]{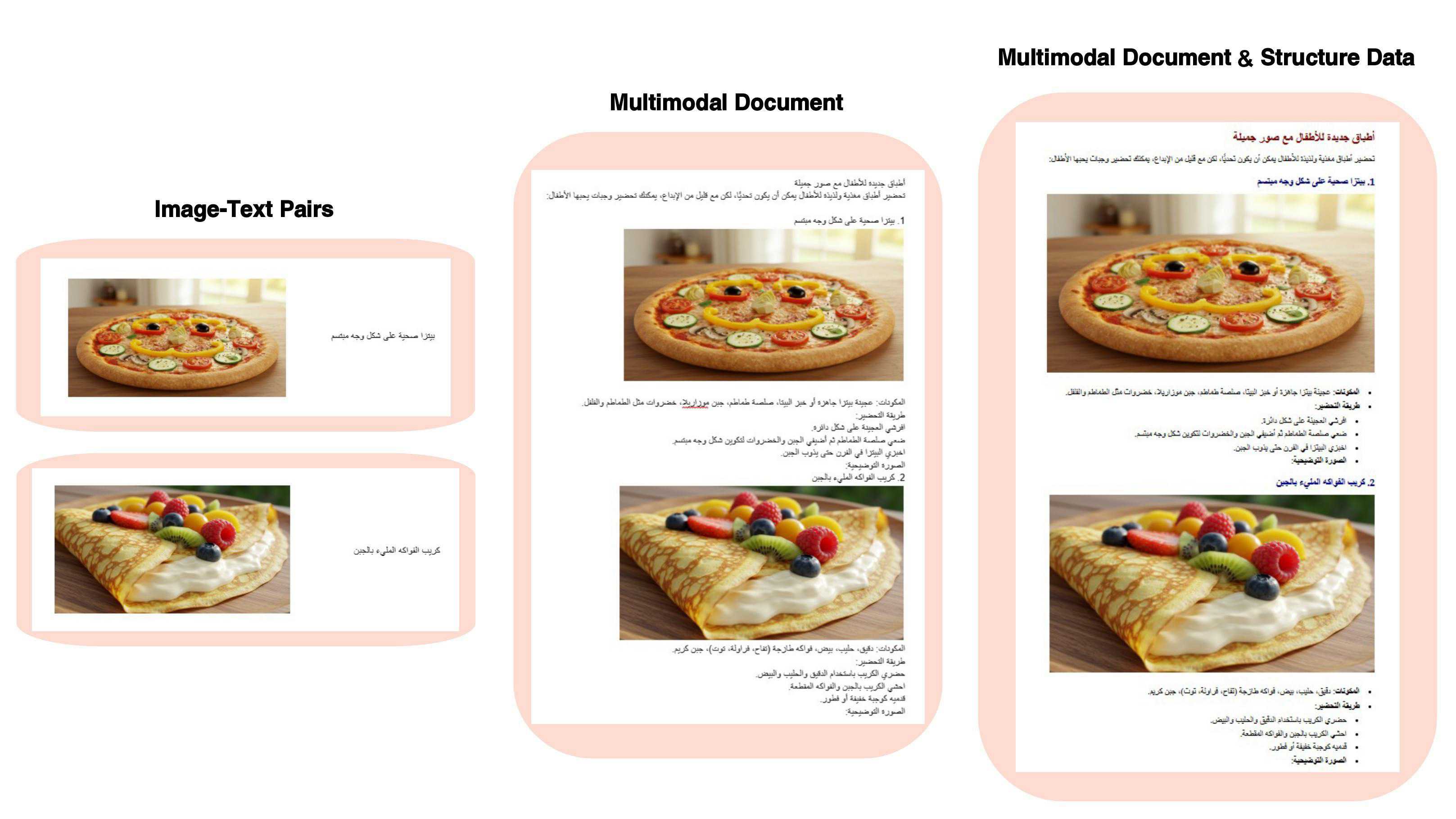}
    \caption{Structured output produced by our pipeline.}
    \label{fig:wasm_result}
\end{figure}

In this paper, we present \textbf{Wasm}, the first Arabic pipeline that produces both interleaved multimodal datasets and text-only corpora with full structural preservation Figure~\ref{fig:pipeline}.
\textit{Wasm} retains document-level structure and the natural interleaving of text and images as they appear on their respective webpage, preserving layout cues and image–text associations needed for multimodal training. Figure~\ref{fig:wasm_result} illustrates the structured output produced by our pipeline. Our work builds upon the OBELICS framework~\cite {OBELICS}, adapting and extending it for Arabic to form an optimized pre-processing pipeline tailored to Arabic web data and multimodal use cases. Unlike OBELICS, which outputs plain text, our approach preserves the structural integrity of web content by converting it into structured Markdown with interleaved images, while maintaining flexibility for both text-only and multimodal pre-training scenarios. Through systematic analysis of filtering techniques, corpus origins, and computational requirements across these datasets, our objective was to establish best practices for Arabic corpus construction and identify the most effective pre-processing strategies for multilingual and multimodal applications.

Our primary contributions are as follows.
\begin{enumerate}
\item We introduce \textbf{Wasm}, a new framework that processes and produces Arabic datasets. To our knowledge, it is the only Arabic pipeline that preserves structural information while maintaining flexibility.
\item We release a part of the Wasm pipeline as open source\footnote{\url{https://github.com/misraj-ai/wasm}}, providing the Arabic-adapted module for interleaved multimodal unformatted data extraction.  This component enables reproducible construction of Arabic web corpora with preserved text–image sequences and can serve as a foundation for further dataset development.
\item We provide a comprehensive comparative analysis of our data processing pipeline against those used for major existing corpora, highlighting the convergences in filtering strategies and justifying our specific design choices.
\item We release a representative dataset dump processed by \textbf{Wasm} \footnote{\url{https://huggingface.co/datasets/Misraj/msdd}}, which, in addition to supporting reproducibility and further research, was used in part to train our vision model, Baseer \cite{hennara2025baseer}.
\end{enumerate}

% \section{Related Work}
\label{sec:related work}

The creation of large-scale, high-quality datasets has been fundamental to recent breakthroughs in large language models (LLMs) and large multimodal models (LMMs). Dataset curation methodologies have evolved from simple crawling approaches to sophisticated multi-stage pipelines that balance scale with quality. This section reviews key developments in the construction of text-only and multimodal datasets, with particular attention to Arabic resources.

\subsection{Text-Only Corpora}

Early large-scale text datasets established foundational principles for web data curation that remain influential today. The ROOTS corpus~\cite{ROOTS}, developed for training BLOOM~\cite{BLOOM}, exemplifies the construction of multilingual datasets by combining diverse sources (Oscar, pseudo-crawls, GitHub) with extensive source-specific filtering and deduplication.

Subsequent work has focused on scaling multilingual coverage while refining quality metrics. CulturaX~\cite{CulturaX}, for example, aggregates the dumps of mC4 and mOSCAR~\cite{moscar} while incorporating perplexity-based scoring, repetition ratios, and confidence thresholds for language identification. These techniques have since become standard in large-scale corpus construction.

FineWeb~\cite{FineWeb2023} represents a recent state-of-the-art corpus. Building on the Massive Text pipeline, it combines neural quality classifiers with custom heuristics and introduces per-dump MinHash deduplication, shown to outperform global deduplication in preserving quality at scale. FineWeb2~\cite{fineweb2} generalizes this pipeline to more than 1000 languages, introducing a scalable multilingual dataset construction framework with language-adaptive filtering and deduplication. In addition, it proposes a principled rebalancing strategy based on duplication and quality metrics, yielding improved downstream model performance across diverse languages.

Arabic-specific curation efforts highlight how pre-processing pipelines adapt to the challenges of non-English web data. The 101 Billion Arabic Words corpus~\cite{101BillionArabic} applied URL filtering, normalization (e.g., Unicode standardization), and document-level deduplication to Common Crawl WET files, while ArabicWeb24~\cite{ArabicWeb24} incorporated more advanced methods such as Gopher quality filtering~\cite{Gopher} and MinHash-based deduplication. However, like all existing Arabic resources, they remain restricted to text-only extraction, discarding structural and multimodal information present in native web documents.

\subsection{Multimodal Corpora}

The evolution toward Vision-Language Models has necessitated datasets that preserve the rich contextual relationships between textual and visual content as they naturally occur on the web. This requirement has driven methodological innovations beyond the simple extraction of image-text pairs.
Early large-scale multimodal datasets, such as LAION-400M~\cite{laion400} and LAION-5B~\cite{laion5b}, were mainly based on large-scale capture of image capture pairs from Common Crawl, filtered using CLIP similarity scores. Although these resources proved invaluable for scaling multimodal training, their pair-based design eliminates broader document context and structural information.

In contrast, more recent work has focused on interleaved multimodal datasets that retain the sequential and structural interplay of text and images within documents. MMC4~\cite{mmc4} advanced beyond basic co-occurrence by incorporating image-text alignment scores and document-level quality metrics into filtering pipelines. This work established the importance of semantic coherence in the construction of multimodal datasets. OBELICS~\cite{OBELICS} marked a further paradigm shift by prioritizing structural preservation: Rather than isolating image-text pairs, OBELICS maintains the interleaved document structure found in web content, creating a corpus of 141 million documents where images and text retain their natural sequential relationships. The processing pipeline operates at both document and HTML node levels, with final output representing documents as coherent sequences of text tokens and contextually-positioned images. Methodologically, OBELICS and MMC4 represent complementary approaches to HTML processing: OBELICS leverages the DOM tree structure for comprehensive content filtering, while MMC4 uses HTML primarily for image location and integration with existing text corpora like C4, highlighting the trade-off between structural fidelity and processing efficiency.

More recently, OmniCorpus~\cite{omnicorpus} has pushed this paradigm further by introducing a 10-billion-level interleaved dataset, incorporating more diverse sources, including English and non-English web domains, as well as video-centric sites, and offers flexible formatting that can be degraded into pure text corpora or image-text pairs. 

MNiT-1T~\cite{mint1T} extends this trajectory by introducing a trillion-token multilingual multimodal corpus constructed from Common Crawl, including new source types such as PDFs and arXiv papers. It combines large-scale image-text pairing with document-level filtering and quality heuristics for robust VLM pre-training.

Complementing these scale-focused efforts, CoMM~\cite{comm} addresses the qualitative limitations of existing interleaved datasets. It introduces a coherent interleaved multimodal corpus, applying multiperspective filters (text coherence, image sequence consistency, image-text alignment) that improve the quality of interleaved training data and enhance the in-context learning capabilities of multimodal LLMs.

To date, Arabic multimodal resources have mainly been based on pairwise translated datasets~\cite{peacock}. None of the major interleaved corpora, such as MMC4, OBELICS, OmniCorpus, MINT-1T, or CoMM, specifically targets Arabic. This gap motivates the development of \textbf{Wasm}, the first Arabic framework to create an interleaved multimodal dataset that preserves structural information and supports both text-only and multimodal pre-training.

% \section{Wasm Pipeline }
\label{sec:pipeline}
This section details the comprehensive data processing pipeline developed to create a structured Arabic dataset. 

\begin{figure}[H]
\centering
\includegraphics[width=\linewidth]{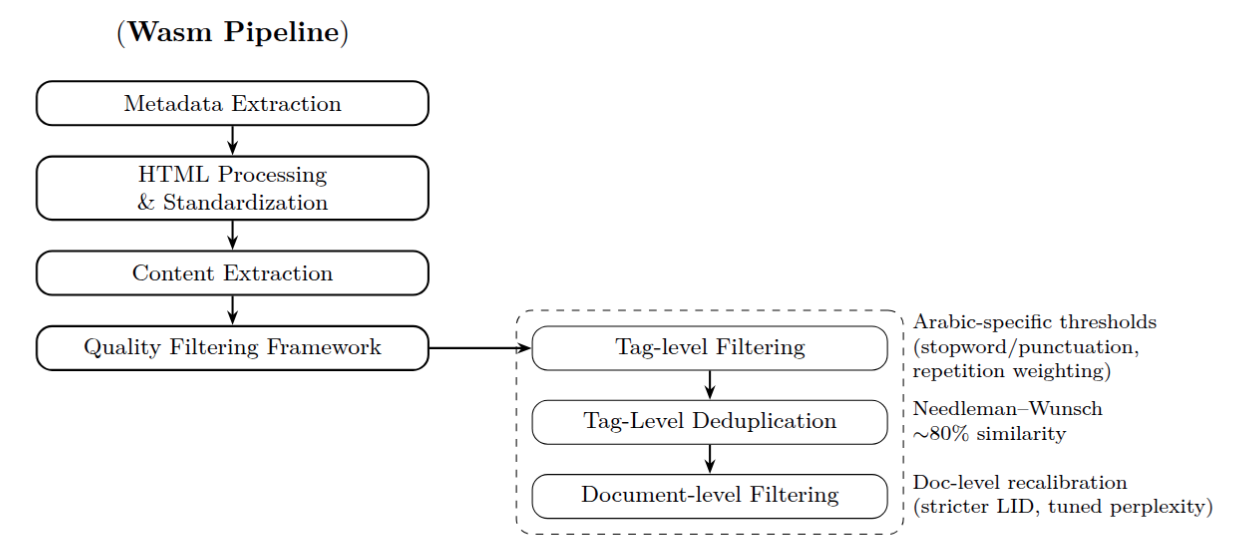}
    \caption{detailed wasm pipline}
    \label{fig:wasm_pipline}
\end{figure}

Our pipeline closely follows the methodology established by OBELICS~\cite{OBELICS}, adapted specifically for Arabic language processing requirements. Figure~\ref{fig:wasm_pipline} provides an overview of the pipeline architecture. All steps in the pipeline were iteratively tuned and refined through careful observation of the outputs on large samples of the data, ensuring that filtering thresholds, normalization rules, and structural preservation techniques were well-suited to the characteristics of Arabic web content.

%%%%%%%%%%
%%%%%%%%%%

\subsection{Metadata Extraction}
\label{subsec:metadata}
The initial phase involves extracting metadata by filtering web pages containing Arabic language content (not necessarily exclusively Arabic) from selected Common Crawl dumps. Each dump was processed separately to facilitate efficient separation and deduplication. Unlike OBELICS, which applies language-based filtering only after loading the Web Archive (WARC) file, our pipeline performs this filtering beforehand. This early intervention not only ensures that irrelevant data is excluded at the source but also saves substantial computational resources, including time, memory, and storage. Each scraped article is then associated with a set of metadata that facilitates storage, retrieval, and analysis. This metadata includes the URL of the article, the storage location within the corresponding Common Crawl snapshot as a WARC file, and the byte position indicating where the webpage begins within that file. In addition, it records the size of the webpage content in bytes, the detected language(s) of the webpage, and the domain name of the source website.

\subsection{HTML Processing and Standardization}
\label{subsec:html_processing}

Using the extracted metadata (filename, offset, and length), the corresponding WARC files were accessed to retrieve the raw webpage content, which was subsequently converted to HTML format. To reduce noise and improve data quality, several preprocessing filters were applied. First, repeated line breaks and whitespace were normalized into single instances to ensure consistency in text formatting. The HTML comments, which do not have semantic value, were then eliminated. The structural elements, such as headers, footers, navigation bars, and menu components, were then removed to retain only the core textual content. Finally, all CSS-related content was removed to eliminate styling artifacts that do not contribute to the linguistic or semantic properties of the data.

\subsection{Simplifying and Structuring Web Content}
\label{subsec:content_extraction}
Following HTML simplification, the content was converted into \textbf{Markdown} format to facilitate downstream text processing, preserve the document's basic structural hierarchy, and allow precise extraction of textual and visual elements. Subsequently, the extraction process categorized the webpage content into two primary types: textual and visual. Unlike the OBELICS pipeline, where text remains largely unstructured, our framework transforms it into structured text; this distinction encompasses elements such as headers, paragraphs, ordered and unordered lists, and tables. The visual category comprises figures (\texttt{fig}) and image (\texttt{img}) tags. To maintain semantic coherence, text elements sharing identical tags are concatenated, thereby preserving both structural integrity and the contextual flow of the content.

\subsection{Quality Filtering Steps}
\label{subsec:filtering}

The pipeline employs a multilevel filtering system at both the tag and document levels to ensure data quality and relevance. A \textit{tag} refers to any connected textual unit (e.g., paragraph, list, or section), while a \textit{document} corresponds to the entire web page.

\subsubsection{Tag-level Filtering}
\label{tag_level_filtering}
Compared to the baseline pipeline \textsc{OBELICS}, we introduced several manual modifications to adapt the filtering process to the specific characteristics of the Arabic language. These adjustments included relaxing or removing certain thresholds that were originally optimized for English but do not generalize well to Arabic.

For example, we reduce the weight of the \textit{Word Repetition Ratio}, which measures the proportion of repeated words in a text. In Arabic, word repetition often carries stylistic and rhetorical significance rather than being indicative of low-quality content. Similarly, we removed the \textit{Stopword Ratio} filter, as Arabic exhibits a rich vocabulary and flexible syntactic structures that allow grammatically correct sentences with relatively few function words. In the same spirit, the \textit{Punctuation Ratio} was also discarded, since Arabic web content frequently lacks punctuation, and using this metric would disproportionately eliminate valid Arabic text.

We also disabled the \textit{Common Word Ratio} filter, which penalizes texts containing many high-frequency words. In Arabic, this would be biased against authentic content, given that the distribution of common words differs significantly from English. Likewise, the \textit{Special Character Ratio} (e.g., emojis, abbreviations) was adjusted more leniently, since contemporary Arabic text often incorporates such elements without necessarily being low-quality.

In contrast, we applied a stricter \textit{Language Identification} process to guarantee that the text is predominantly Arabic, while still allowing for the natural occurrence of foreign terms (for example, English terminology). Recognizing that the original perplexity model did not meet our requirements, we developed a customized version based on the KenLM~\cite{kenlm} framework. This model was trained on a carefully curated dataset that emphasizes high-quality content and spans a wide spectrum of Arabic dialects and topics. This diversity enhances the robustness of our filtering pipeline, ensuring that the retained text is both representative and linguistically rich. Finally, the \textit{Perplexity Threshold} was meticulously calibrated to eliminate incoherent or machine-generated text (e.g., spam, low-quality advertisements, or poorly generated AI output), while preserving the integrity of well-formed human-authored Arabic across dialectal and topical variations.

It should be noted that Arabic remains a low-resource language on the Web, representing only about 0.6\% of the content in the Common Crawl datasets \cite{commoncrawl}. Overly restrictive thresholds would therefore risk discarding a substantial amount of the already scarce Arabic data.

\subsubsection{Visual Data Filtering}
Our image data filtering strategy was tailored to the characteristics of Arabic web content, with an emphasis on maximizing data retention while upholding quality standards. Given the relative scarcity of Arabic multimodal resources, we adopted a conservative approach that avoids unnecessary exclusions. 

Instead of downloading images directly, we collected their URLs to reduce storage costs, accelerate the acquisition process, and enable scalable filtering. This design naturally shifted the focus of filtering from individual images to the URL level. To ensure safety and appropriateness, we maintain a blacklist of websites that host explicit or unsuitable material and exclude all associated image URLs. In particular, Arabic web content is rarely hosted on mainstream platforms that contain prohibited content, which further justifies this site-level filtering strategy.

The resulting set of URLs forms a flexible foundation for subsequent stages of image processing and task-specific filtering, allowing later adjustments to be aligned with the requirements of different models and training objectives.

\subsubsection{Tag-Level Deduplication}

We have worked to remove implicit duplicates within documents. For example, some sites contain duplicate ads. Unlike the standard OBELICS pipeline, which would reject an entire document for such duplication, we avoid full deletion whenever possible.

To address substantial repetition at the tag level observed in the dataset, we implemented the Needleman–Wunsch algorithm~\cite{Needleman} with a similarity threshold of 80\% to efficiently identify and remove nearly duplicate content.

\subsubsection{Document-level Filtering}
At the document level, we applied the same set of filtering criteria as in the tag-level filtering \ref{tag_level_filtering}, but with different parameter values. These values were recalibrated to reflect document-wide characteristics rather than paragraph-level ones. In particular, thresholds were tuned to balance the need for higher-quality long-form content with the goal of retaining as much Arabic data as possible.

% As shown in Appendix~\ref{app:filter_table}, our Arabic-focused filtering introduces several modifications and adaptations.

% \section{Discussion}
\label{sec:discussion}

This section analyzes our methodological contributions in the context of existing dataset construction approaches, examining how our design choices address key limitations identified in previous work while advancing the state-of-the-art in Arabic multimodal dataset curation. Our design strategies are explained in Table \ref{tab:method_differences}

\subsection{Methodological Innovations and Comparative Analysis}

Our approach introduces three fundamental improvements over existing methodologies that collectively enhance both dataset quality and structural fidelity.

\textbf{Structured Data Preservation.} Unlike approaches that flatten web content into sequential text-image pairs (e.g., MMC4~\cite{mmc4}) or transform documents into linear token sequences (e.g., OBELICS~\cite{OBELICS}), our methodology preserves the hierarchical structure inherent in web documents in Markdown format. This preservation maintains semantic relationships between content elements such as image-caption associations, section hierarchies, and contextual dependencies that are crucial for training models capable of understanding document-level coherence. Although OBELICS maintains an interleaved structure, our approach goes further by preserving the underlying DOM hierarchy, enabling more sophisticated downstream applications that require an understanding of document organization and content relationships.

\textbf{Enhanced Perplexity-Based Quality Assessment.}
Expanding the perplexity-based filtering strategy proposed in OBELICS~\cite{OBELICS}, where a KenLM model was trained on Wikipedia to evaluate text quality, we refined the approach to better detect and remove incoherent or automatically generated material. Our method places a greater emphasis on safeguarding the authenticity of human-produced Arabic text, capturing both dialectal richness and topical breadth. To this end, the model was trained on a carefully balanced corpus that prioritizes linguistic fidelity and diversity, ensuring representation across multiple Arabic dialects and subject areas while maintaining consistently high standards of quality.

The performance of our model was systematically compared with that of a counterpart trained solely on Arabic Wikipedia KenLM~\footnote{\url{https://huggingface.co/edugp/kenlm}} to determine its filtering effectiveness. Empirical evaluation revealed consistently superior filtering performance by our model in an extensive suite of examples, indicating significant deficiencies in its quality control mechanisms (see Table~\ref{tab:perplexity_comparison}).

To assess the performance of our model, we evaluated multiple datasets. 
For each dataset, we randomly sampled 100,000 examples and calculated the perplexity for each instance. 
Based on these calculations, we determined the \textit{exclusion rate}, defined as the proportion of examples rejected by the model due to the exceeding of acceptable perplexity thresholds. 
Table~\ref{tab:perplexity_exclusion} reports the exclusion rates for each dataset. 
To provide qualitative insight into the nature of the excluded data, 
representative examples are presented separately in Table~\ref{tab:perplexity_examples}.
This separation allows for a clearer distinction between the quantitative summary and the qualitative illustration of the model's filtering behavior.

\begin{table}[ht]
\centering
\caption{Exclusion rates across datasets based on perplexity thresholds.}
\label{tab:perplexity_exclusion}
\resizebox{0.4\linewidth}{!}{%
\begin{tabular}{l c}
\toprule
\textbf{Dataset} & \textbf{Exclusion Rate (\%)} \\
\midrule
Wasm & 0 \\
fine_web2 & 1.766 \\
ara24 & 7.82 \\
cultura_x & 8.605  \\
dataset_101 & 19.757 \\
\bottomrule
\end{tabular}
}
\end{table}

\begin{table}[ht]
    \centering
    \begin{tabular}{c}
     \includegraphics[
     clip,
     % left bottom right top
     trim=0cm 14.6cm 0cm 2cm, 
     width=1\textwidth]{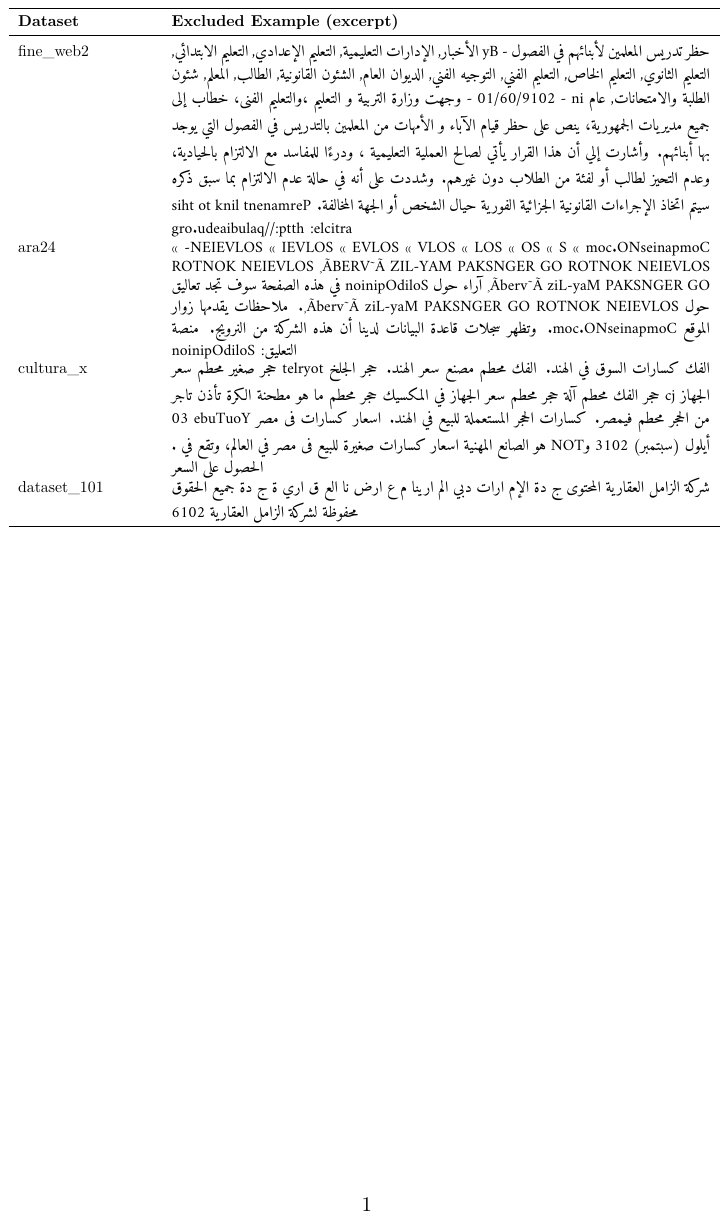}\\
    \end{tabular}
    \caption{Representative examples of excluded data based on high perplexity values.}
\label{tab:perplexity_examples}
\end{table}

\textbf{Granular Node-Level Deduplication.} While existing approaches typically perform deduplication at the document level (e.g., ROOTS~\cite{ROOTS}, 101 Billion Arabic Words~\cite{101BillionArabic}) or apply MinHash deduplication globally or per-dump (e.g., FineWeb~\cite{FineWeb2023}), our methodology implements deduplication at the HTML node level. This granular approach enables the preservation of documents that contain unique content alongside duplicated elements (such as navigation menus or boilerplate text), significantly improving content diversity while maintaining processing efficiency. Node-level deduplication is particularly valuable for web documents where substantial unique content may coexist with repeated structural elements, allowing for more nuanced quality preservation than binary document-level decisions.

\begin{table}[ht]
\centering
\caption{Key methodological differences between OBELICS and Wasm and their impact on dataset utility.}
\label{tab:method_differences}
\begin{tabularx}{\textwidth}{|X|X|X|X|}
\hline
\textbf{Aspect} & \textbf{OBELICS} & \textbf{Wasm (Ours)} & \textbf{Impact / Motivation} \\
\hline
Quality Filtering & Aggressive filtering with multiple constraints & Balanced filtering adapted to Arabic & 
Maintains high quality while preserving Arabic linguistic structures \\
\hline
Document Structure & Sequential interleaved format & Preserved structure with separate columns & Facilitates extraction of both text and visual content for multimodal model training \\
\hline
Perplexity Assessment & Limited use with English Wikipedia-based KenLM & Central criterion with Arabic-tuned thresholds across dialects & Retains valid Arabic variation while filtering incoherent or low-quality text \\
\hline
Deduplication Strategy & \centering
N/A & Sequence alignment-based & More accurate removal of near-duplicate Arabic content \\
\hline
Content Flexibility & Optimized for specific data type & Flexible for diverse data types and tasks & Supports training of multiple model types and tasks \\
\hline
\end{tabularx}
\end{table}

\section{Conclusion}
\label{sec:conclusion}

This study introduces \textsc{Wasm}, the first large-scale Arabic multimodal processing framework built on Common Crawl data, designed to preserve the structural and semantic integrity of web documents, including the natural interleaving of text and images. Unlike prior text-only efforts, \textsc{Wasm} provides a flexible foundation for training both LMMs and LLMs by maintaining document-level coherence, cross-modal alignments, and hierarchical structures such as captions, sections, and contextual dependencies. The framework integrates Arabic-specific perplexity modeling, dialectal coverage, and KenLM-based adaptive filtering to ensure linguistic fidelity, alongside fine-grained node-level deduplication using Needleman–Wunsch, achieving higher corpus diversity and efficiency than conventional document-level approaches. By releasing both the dataset and pipeline code, \textsc{Wasm} not only democratizes access to advanced multimodal Arabic resources but also pushes the boundaries of Arabic NLP development, enabling reproducible research and laying the groundwork for future large-scale corpus construction.

\bibliography{main}

\clearpage
\appendix
\section{Filtering Parameters Comparison}
\label{app:filter_table}

The following table provides a detailed comparison of the filtering parameters of OBELICS with our Arabic-focused adaptations.  

\begin{table}[ht]
    \centering
    \begin{tabular}{c}
     \includegraphics[
     clip,
     % left bottom right top
     trim=0cm 8.6cm 0cm 2cm, 
     width=1\textwidth]{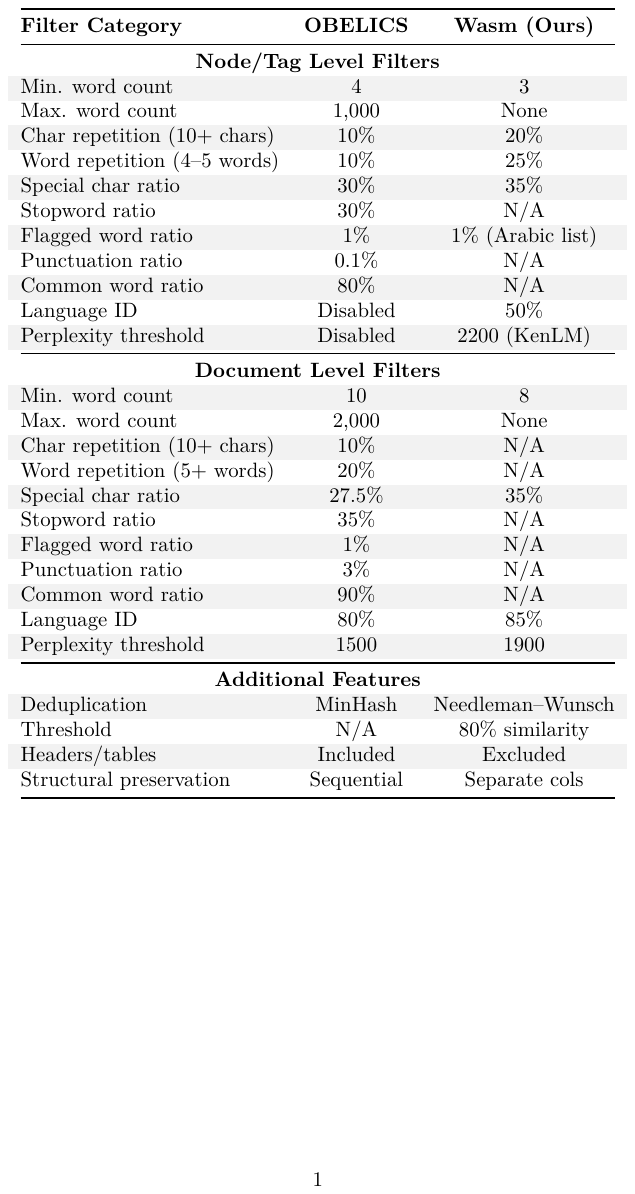}\\
    \end{tabular}
    \caption{Filtering Parameters Comparison}
\label{tab:filter_comparison}
\end{table}

\clearpage

\section{Perplexity Model Comparison}
\begin{table}[ht]
    \centering
    \begin{tabular}{c}
     \includegraphics[
     clip,
     % left bottom right top
     trim=0cm 8.6cm 0cm 2cm, 
     width=1\textwidth]{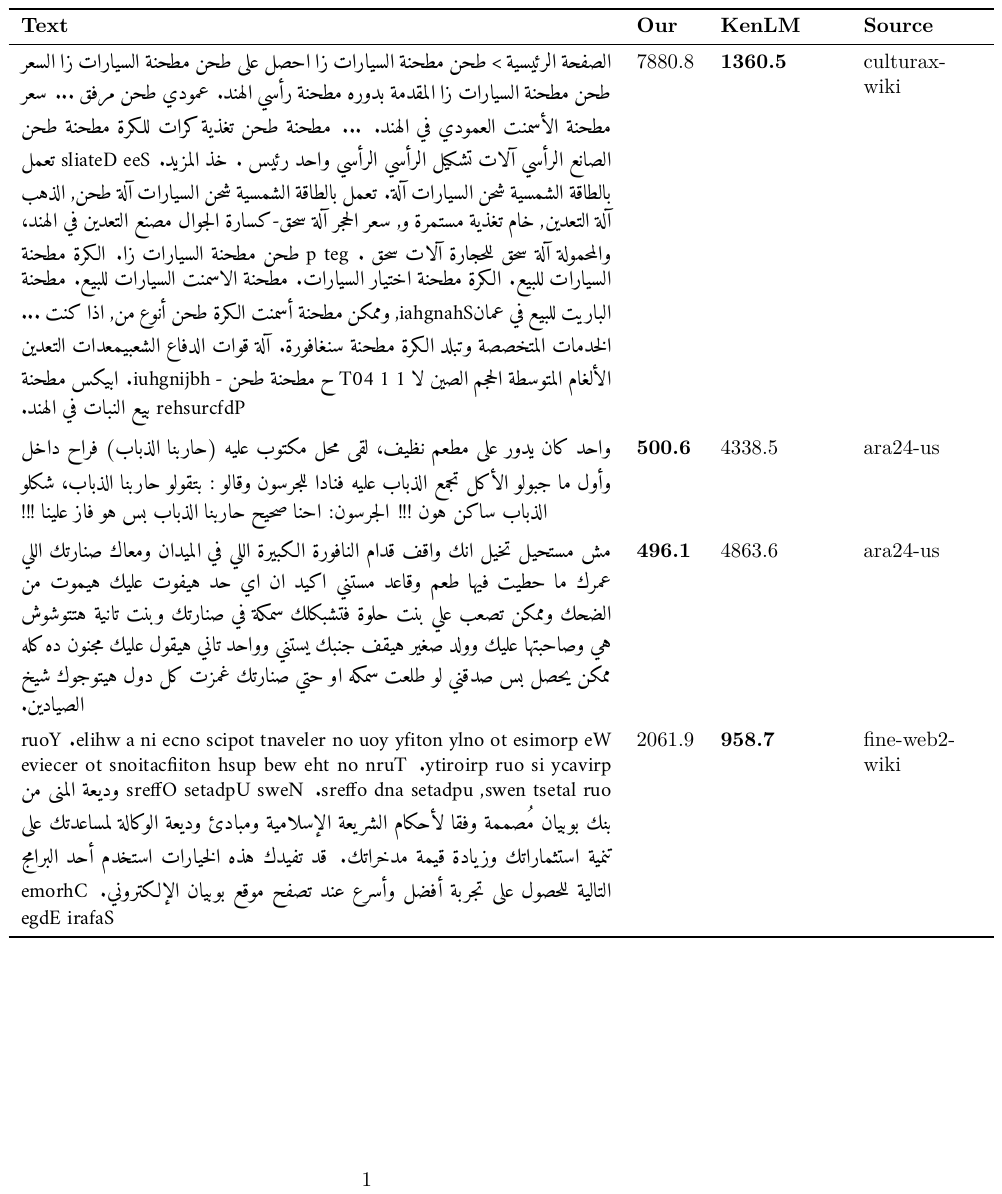}\\
    \end{tabular}
    \caption{Perplexity comparison on sample Arabic text. Values in \textbf{bold} indicate the lower (better) perplexity for that sample. This table highlights examples where the KenLM/Wikipedia model gives a much lower perplexity than our model (a potential issue that requires inspection).}
\label{tab:perplexity_comparison}
\end{table}

\end{document}